\documentclass[conference]{IEEEtran}
\IEEEoverridecommandlockouts
\usepackage{cite}
\usepackage{amsmath,amssymb,amsfonts}
\usepackage{algorithmic}
\usepackage{graphicx}
\usepackage{textcomp}
\usepackage{xcolor}
\usepackage{arydshln}

\def\BibTeX{{\rm B\kern-.05em{\sc i\kern-.025em b}\kern-.08em
    T\kern-.1667em\lower.7ex\hbox{E}\kern-.125emX}}
\begin{document}

\title{Competitive Ensembling Teacher-Student Framework for Semi-Supervised Left Atrium MRI Segmentation\\

}

\author{\IEEEauthorblockN{Yuyan Shi}
\IEEEauthorblockA{\textit{Network and Data Center} \\
\textit{Northwest University}\\
Xi’an, China \\
shiyuyan@stumail.nwu.edu.cn}
\and
\IEEEauthorblockN{Yichi Zhang}
\IEEEauthorblockA{\textit{School of Data Science} \\
\textit{Fudan University}\\
Shanghai, China \\
yichizhang.coda@gmail.com}
\and
\IEEEauthorblockN{Shasha Wang}
\IEEEauthorblockA{\textit{Network and Data Center} \\
\textit{Northwest University}\\
Xi’an, China \\
wss@nwu.edu.cn}}

\maketitle

\begin{abstract}
Semi-supervised learning has greatly advanced medical image segmentation since it effectively alleviates the need of acquiring abundant annotations from experts and utilizes unlabeled data which is much easier to acquire. Among existing perturbed consistency learning methods, mean-teacher model serves as a standard baseline for semi-supervised medical image segmentation.
In this paper, we present a simple yet efficient competitive ensembling teacher student framework for semi-supervised for left atrium segmentation from 3D MR images, in which two student models with different task-level disturbances are introduced to learn mutually, while a competitive ensembling strategy is performed to ensemble more reliable information to teacher model.
Different from the one-way transfer between teacher and student models, our framework facilitates the collaborative learning procedure of different student models with the guidance of teacher model and motivates different training networks for a competitive learning and ensembling procedure to achieve better performance.
We evaluate our proposed method on the public Left Atrium (LA) dataset and it obtains impressive performance gains by exploiting the unlabeled data effectively and outperforms several existing semi-supervised methods.
\end{abstract}

\begin{IEEEkeywords}
Semi-supervised learning, Medical Image Segmentation, Competitive Ensembling
\end{IEEEkeywords}

\section{Introduction}

Left atrium is a vital cardiac structure responsible for receiving oxygenated blood from the pulmonary veins and delivering it to the left ventricle for systemic circulation.
Accurate segmentation of the left atrium from cardiac MRI images can provide essential anatomical insights and quantitative measurements that are invaluable for diagnosing and monitoring various cardiovascular conditions, which plays a crucial role in clinical practice \cite{xiong2021global}. 

Manual segmentation of the left atrium is a tedious and labor-intensive task that is prone to inconsistencies between different observers. Therefore, there is a growing need to develop automatic segmentation methods.
In recent years, deep learning techniques have demonstrated remarkable advancements and achieved state-of-the-art results in various medical image segmentation tasks \cite{ma2021abdomenct,lalande2021deep}. However, the success of training deep neural networks heavily relies on a large amount of labeled datasets, while obtaining such datasets for is both expensive and time-consuming, especially for medical imaging where reliable annotations can only be provided by professional experts \cite{tajbakhsh2020embracing}.

\begin{figure*}[t]
	\includegraphics[width=18cm]{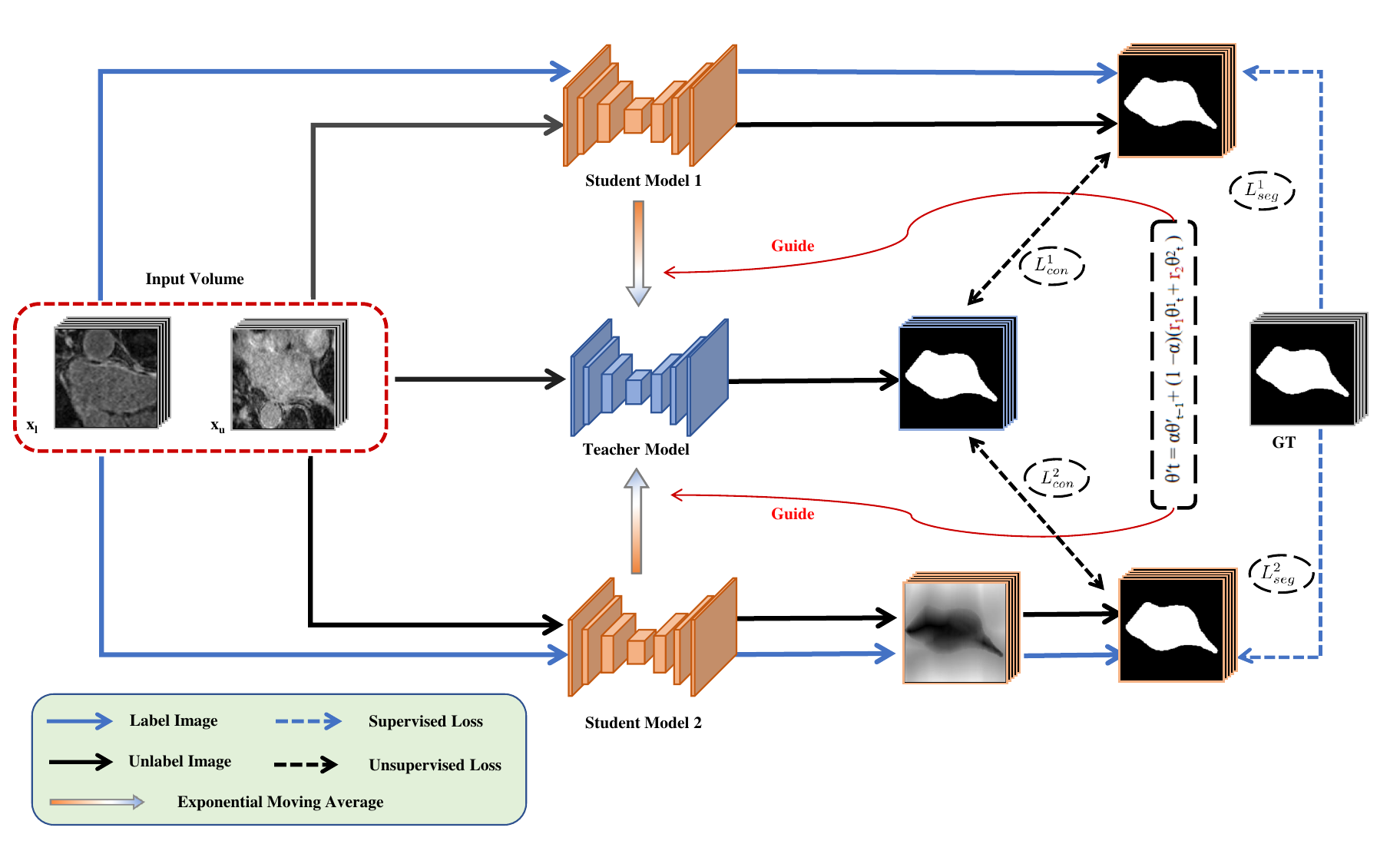}
	\caption{The architecture of our proposed competitive ensembling framework with comparison to classic self-ensembling framework.}
	\label{Fig1}
\end{figure*}

To minimize the high cost associated with labeling medical images, numerous studies have been dedicated to developing annotation-efficient methods for medical image segmentation with limited annotations \cite{zhang2021exploiting,fang2022annotation}.
Among these approaches, semi-supervised segmentation is a more practical method by encouraging segmentation models to learn with limited amount of labeled data and large amount of unlabeled data which is much easier to acquire \cite{SemiSurvey}.
As one of the most important baseline approaches for semi-supervised learning, the mean teacher model \cite{tarvainen2017mean} has achieved great success by enforcing the consistency of predictions from perturbed inputs between student and teacher models.
Based on this, several teacher-student framework-based methods \cite{yu2019uncertainty,wang2021tripled,zhou2022dtsc,zhang2023uncertainty} have been proposed for semi-supervised medical image segmentation.
However, the classic one-way transfer between teacher and student models may introduce unsatisfying performance due to the limited annotations.

In this paper, we propose a novel competitive ensembling teacher student framework for semi-supervised for left atrium segmentation from 3D MR images. Specifically, two student models with different task-level disturbances are introduced to learn mutually from different perspectives. Instead of classic self-ensembling, we implement a competitive ensembling strategy to combine more reliable information to teacher model.
Different from the one-way transfer between teacher and student models, our framework promotes the collaborative learning procedure of different student models under the guidance of teacher model and encourages different training networks to conduct a competitive learning and integration process. Ultimately leading to achieve better performance.
We evaluate our proposed method on the public Left Atrium (LA) dataset and it obtains impressive performance gains by exploiting the unlabeled data effectively and outperforms several existing semi-supervised methods.

\section{Related Work}

\subsection{Medical Image Segmentation}

The purpose of image segmentation is to understand the image at the pixel level and classify each pixel into a certain class. Recently, convolutional neural network has achieved advanced performance in many medical image segmentation tasks.
The most popular structure for medical image segmentation is U-Net \cite{ronneberger2015u} which utilizes an encoder-decoder architecture to fuse features of different scales with skip connections. 
For volumetric medical images, 3D segmentation networks such as 3D U-Net \cite{cciccek20163d} and V-Net \cite{milletari2016v} are proposed to use 3D convolutional kernels to extract volumetric features. In addition, many variants of U-Net have been proposed to improve it by designing new structures and have been applied to many medical image segmentation tasks \cite{zhang2017deep,li2018h,zhang2020sau}.

\subsection{Semi-Supervised Learning}

To alleviate the heavy burden of acquiring annotations, researchers have focused on developing semi-supervised segmentation methods that leverage limited amount of labeled data along with large amount of unlabeled data to reduce the overall manual cost while maintaining the segmentation performance.
There are two main types of semi-supervised learning methods. The first type is based on pseudo labels, where unlabeled images are assigned pseudo annotations and used to update the segmentation model. For example, Bai et al. \cite{bai2017semi} iteratively updated pseudo segmented labels and network parameters, and refined the pseudo labels using Conditional Random Fields (CRF). Zhang et al. \cite{zhang2017deep} introduced adversarial learning for biomedical image segmentation, which aims to encourage the segmentation output of unlabeled data to be similar to the annotations of labeled data. 
Another category of semi-supervised learning aims to learn from both labeled and unlabeled images by utilizing a combination of supervised loss for labeled images and unsupervised loss for unlabeled or all images. Among these methods, consistency learning is widely employed to enhance the invariance of predicted image outputs under different distributions.
Li et al. \cite{li2020transformation} applied perturbations like Gaussian noise, randomly rotation
and scaling to the input images and encourage the network to be transformation-consistent for unlabeled data. Yu et al. \cite{yu2019uncertainty} extended the mean teacher paradigm with an uncertainty estimation strategy to improve the performance of consistency-based model so as to learn from more meaningful and reliable targets during training.

\section{Method}

In this work, we introduce a novel competitive ensembling mean teacher framework (CE-MT) to improve classic mean teacher model for semi-supervised medical image segmentation.
An overview of our proposed method is shown in Fig \ref{Fig1}. Different from classic mean teacher framework, our framework consists of a teacher model and two student models with different learning conditions. The student models are trained to minimize both the supervised loss on the labeled data and the consistency loss with respect to the output from the teacher model.
Different from the classic one-way transfer between teacher and student models, a competitive ensembling strategy is proposed to select out more reliable information from student models for the exponential moving average to update the teacher model.

\subsection{Dual Task Student Models} \label{network}

As an alternative to classic pixel-level classification for image segmentation tasks, distance maps aim to generate a gray-level image where the intensities of pixels are changed according to the distance to the closest boundary, which have been incorporated to several medical image segmentation tasks \cite{zhang2021dual,zhang2023uncertainty}.

Based to the design of encoder-decoder architecture \cite{ronneberger2015u,cciccek20163d,milletari2016v}, other than classic segmentation head to generate probabilistic maps, an auxiliary regression head composed by a 3D convolution block followed by the \textit{tanh} activation is added to generate the distance maps as follows:
 
\begin{equation}
G_{SDF}=\left\{
\begin{array}{lllll}
-\inf \limits_{y \in \partial G}\|x-y\|_{2} , & x \in G_{\mathrm{in}}  \\
\\
 0, & x \in \partial G \\
 \\
+\inf \limits_{y \in \partial G}\|x-y\|_{2} , & x \in G_{\mathrm{out}}  \\
\end{array}\right.
\end{equation}
where $\|x-y\|_{2}$ is the Euclidean distance between voxels $x$ and $y$, and $G_{in}$, $\partial G$, $G_{out}$ represent the inside, boundary and outside of the target object. $G_{SDF}$ takes negative values inside the target and positive values outside the object, and the absolute value is defined by the distance to the closest boundary point.
To transform the output of signed distance maps to segmentation output, we utilize a smooth approximation to the inverse transform which can be defined by

\begin{equation} \label{trans}
G_{mask}=\frac{1}{1+e^{-k \cdot z}} , z \in G_{SDF}
\end{equation}

In our framework, two student networks share the same backbone structure and differ at the output branch of the network. Specifically, segmentation student model $M_{1}$ only activate the segmentation head, while regression student model $M_{2}$ only activate the regression head.

\subsection{Mean Teacher Model}

Regarding the semi-supervised segmentation task, we denote the labeled set as $D_{L} = \{X_{i},Y_{i}\}_{i=1}^{M}$ and the unlabeled set as $D_{U} = \{Y_{i}\}_{i=1}^{N}$, where $X_{i} \in R^{H \times W \times D}$ is the input medical images and $Y_{i} \in \{0,1\}^{H \times W \times D}$ is corresponding ground truth. 
In mean teacher framework, the teacher model is an average of student model over different training steps to produce a more accurate model than using the final weights directly. Instead of sharing the weights with the student model, the teacher model uses exponential moving average (EMA) to ensemble weights of the student model as $ \theta^{'}_{t} = \alpha \theta^{'}_{t-1} + (1-\alpha) \theta_{t}$, where $\theta$ and $ \theta^{'}$ represent the weights of student model and teacher model, and $\alpha$ is a hyper-parameter named EMA decay.

For semi-supervised learning, the student model learns from the teacher model by minimizing the combination of supervised segmentation loss and unsupervised consistency loss with respect to the output targets of the teacher model. 
Therefore, the task can be formulated as training the network by minimizing the following functions.

\begin{equation}
\min \limits_{\theta} \mathcal{L}_{sup}(\theta;\mathcal{D}_{L}) + \lambda_{i} \mathcal{L}_{con}(\theta;\theta^{'};\mathcal{D})
\end{equation}

\subsection{Competitive Ensembling Teacher-Student Framework}

However, the traditional one-way transfer of knowledge from teacher to student models may yield sub-optimal performance due to the scarcity of annotations.
To achieve better training, we introduce a novel competitive ensembling strategy to select out more reliable information from different student models for the exponential moving average (EMA) to update the teacher model.
Different from classic mean teacher framework, our competitive ensembling framework consists of a teacher model and two student models with different learning conditions as illustrated in section \ref{network}. Specifically, segmentation student model named $M_{1}$ only activate the segmentation head, while regression student model named $M_{2}$ only activate the regression head.

For training, the student models is optimized by minimizing a combination of the supervised loss on the labeled data and the consistency loss with respect to the output from the teacher model. This minimization is performed with respect to the output targets provided by the teacher model.

\begin{equation}
\min \limits_{\theta^{1}} \mathcal{L}_{sup}(\theta^{1};\mathcal{D}_{L}) + \lambda_{con} \mathcal{L}_{con}(\theta^{1};\theta^{'};\mathcal{D})
\end{equation}

\begin{equation}
\min \limits_{\theta^{2}} \mathcal{L}_{sup}(\theta^{2};\mathcal{D}_{L}) + \lambda_{con} \mathcal{L}_{con}(\theta^{2};\theta^{'};\mathcal{D})
\end{equation}
where $\theta^{1}$ and $\theta^{2}$ represent the weights of segmentation student model $M_{1}$ and regression student model $M_{2}$, and $\theta^{'}$ represents the weights of teacher model $M_{t}$.
For supervised loss, we employ the combination of dice loss and cross-entropy loss as the supervised loss $\mathcal{L}_{sup}$ for the segmentation.

Different from the classic one-way transfer between teacher and student models, a competitive ensembling strategy is proposed to select out more reliable information from student models for the exponential moving average to update the teacher model, which can be formulated as follows:

\begin{equation}
\theta_{t}^{'} = \alpha\theta_{t-1}^{'} + (1-\alpha)(r_{1}\theta_{t}^{1} + r_{2}\theta_{t}^{2} )
\end{equation}
where $r_{1}$ and $r_{2}$ are competitive parameters for the ensembling of different student models.
Here we utilize the supervised dice loss (i.e. segmentation performance) of the two student networks on labeled dataset for the selection of $r_{1}$ and $r_{2}$ to decide to ensemble which student model to the teacher model.
Specifically, two different competitive ensembling strategies are adopted to update the teacher model, which will be introduced in the following paragraphs.

1) \textbf{Unidirectional Competitive Ensembling  (CE-MT-U).} 
To select out more reliable information from different student models, one simple and directive method is using the student model with "better performance" for the exponential moving average (EMA) to update the teacher model, while the other one is temporarily neglected. 
For better performance, we utilize the supervised loss function of the two student networks on labeled dataset for the selection to decide to ensemble which student model to the teacher model.

For this straightforward and simple strategy, we compare the loss function directly between the two student models. If the supervised dice loss of $M_{1}$ is lower than that of  $M_{2}$, the parameter of student model $M_{1}$ is used for ensembling with the teacher model. Conversely, if the supervised dice loss of  $M_{1}$ is higher, the parameter of student model $M_{2}$ is selected for ensembling with the teacher model. 
The definition of competitive parameters $r_{1}$ and $r_{2}$ are shown as follows:

\begin{align}
r_{1} = \mathbb{I}( \mathcal{L}_{dice}(\theta^{1};\mathcal{D}_{L}) \textless  \mathcal{L}_{dice}(\theta^{2};\mathcal{D}_{L})) \notag \\
r_{2} = \mathbb{I}( \mathcal{L}_{dice}(\theta^{1};\mathcal{D}_{L}) \geq  \mathcal{L}_{dice}(\theta^{2};\mathcal{D}_{L}))
\end{align}

2) \textbf{Bidirectional Competitive Ensembling (CE-MT-B).} 
The limitation of previously introduced strategy is that if one student model's learning effect is significantly better than the other student model in the early training procedure, the teacher model tends to integrate from the superior student model. As a result, the student model may exhibit higher consistency with the teacher model, further exacerbating this difference. This outcome could lead to dominant learning from only one student model and causing the framework to degrade into classic mean teacher framework.


Therefore, we introduce another strategy where both two student models are ensembled to the teacher model with different competitive parameters $r_{1}$ and $r_{2}$. If the supervised dice loss of $M_{1}$ is lower than that of $M_{2}$, the competitive parameters $r_{1}$ is higher than $r_{2}$, and vise versa. In this way, the information of different tasks with different student models can be utilized in an competitive learning manner.
The definition of competitive parameters $r_{1}$ and $r_{2}$ are shown as follows:

\begin{align}
r_{1} = \frac{1-{L}_{dice}(\theta^{1};\mathcal{D}_{L})} {2 - {L}_{dice}(\theta^{1};\mathcal{D}_{L}) - {L}_{dice}(\theta^{2};\mathcal{D}_{L})} \notag \\
r_{2} = \frac{1-{L}_{dice}(\theta^{2};\mathcal{D}_{L})} {2 - {L}_{dice}(\theta^{1};\mathcal{D}_{L}) - {L}_{dice}(\theta^{2};\mathcal{D}_{L})}
\end{align}


For the update competitive ensembling strategy, the teacher model no longer only integrates one of the two student models. Instead, the framework aims to enhance the integration of useful information while reducing the incorporation of ineffective information by employing the competitive parameters $r_{1}$ and $r_{2}$, ensuring that the knowledge acquired from each task is given due consideration and not overlooked.

\begin{table}[t]
	\caption{Experimental results of ablation analysis of our proposed competitive ensembling teacher student framework under different settings.} \label{Table0}
	\centering
	\renewcommand\arraystretch{1.5}
	\begin{tabular}{c|c|c|c|c|c}
		\hline 
		\textbf{Method}	& \textbf{Label/Unlabel}	& \textbf{Dice[\%]} & \textbf{Jaccard[\%]} & \textbf{ASD} & \textbf{95HD}  \\ \hline
        MT  \cite{tarvainen2017mean}   & 8/72             & 82.38 & 72.32  & 2.21 & 14.54    \\
            \textbf{CE-MT(u)} & 8/72        & 87.63 & 78.18 & 1.99 & 9.56  \\  
            \textbf{CE-MT(b)} & 8/72     & 87.96 & 78.76 & 1.99 & 8.26  \\\hline 
            MT \cite{tarvainen2017mean}    & 16/64             & 88.23     & 79.29       & 2.73        & 10.64     \\
		\textbf{CE-MT(u)} & 16/64        & 89.77 & 81.61 & 1.80 & 6.78  \\  
            \textbf{CE-MT(b)} & 16/64        & 89.78 & 81.62 & 1.80 & 6.78  \\  \hline 
	\end{tabular}
\end{table}

\begin{table*}[h]
	\caption{Comparative experimental results between our proposed method and other semi-supervised segmentation methods on LA dataset with 20\% annotation settings (16 labeled scans and 64 unlabeled scans).} \label{Table1}
	\centering
	\renewcommand\arraystretch{1.5}
	\begin{tabular}{c|c|c|c|c|c}
		\hline 
		\textbf{Method}	& \textbf{Label/Unlabel}	& \textbf{Dice [\%]} & \textbf{Jaccard [\%]} & \textbf{ASD} & \textbf{95HD}  \\ \hline
		Supervised Baseline &  16/0              & 86.03     & 76.06        & 3.51        & 14.26        \\ \hline
        DAN \cite{zhang2017deep}    & 16/64                  &87.52   &78.29        &2.42   &9.01        \\        
        CPS  \cite{chen2021semi}   &16/64                 &87.87    &78.61 &2.16   &12.87               \\
        Entropy Mini \cite{vu2019advent}   &16/64           &88.45   &79.51  &3.90   &14.14                       \\ 
        DUW \cite{wang2020double}  &16/64                   &89.65 &81.35  &2.03  &7.04                        \\
        ICT \cite{verma2022interpolation}  &16/64                &89.02  &80.34 &1.97 &10.38               \\
        SASSNet \cite{li2020shape}   &16/64                &89.27  &80.82   &3.13  &8.83                   \\
        DTC  \cite{luo2021semi}  &16/64                      &89.42   &80.98   &2.10   &7.32              \\
        CCT \cite{ouali2020semi}   &16/64                   &88.01   &80.95    &2.37    &8.25                  \\
        URPC \cite{luo2022semi}  &16/64                   &88.43   &81.15   &2.35    &8.21                     \\
        CPCL \cite{xu2022all} &16/64                         &88.32   &81.02   &2.02   &8.01              \\ \hline
        MT \cite{tarvainen2017mean}    & 16/64             & 88.23     & 79.29       & 2.73        & 10.64     \\
	UA-MT  \cite{yu2019uncertainty}   & 16/64             & 88.88     & 80.21       & 2.26        & 7.23     \\
	\textbf{CE-MT(ours)} & 16/64      & \textbf{89.78} & \textbf{81.62} & \textbf{1.80} & \textbf{6.78}  \\   \hline 
	\end{tabular}
\end{table*}

\begin{figure*}[t]
	\includegraphics[width=18cm]{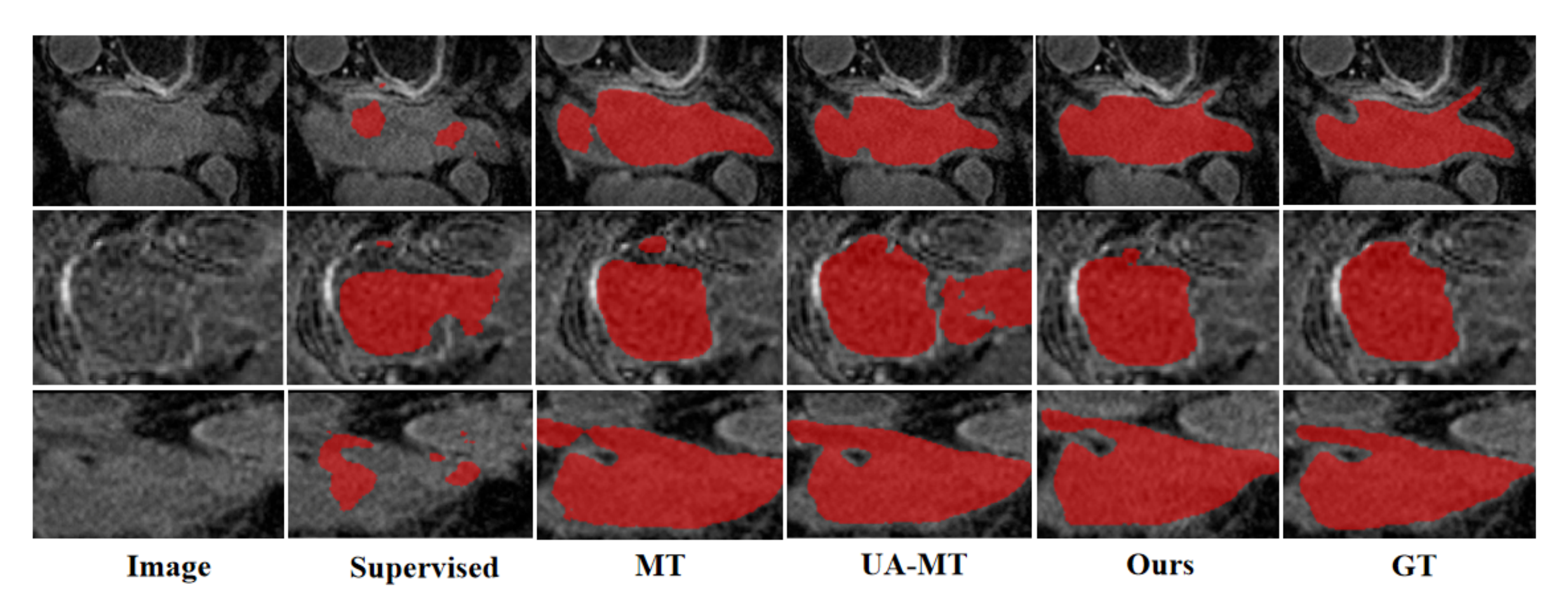}
	\caption{Visual comparison of segmentation results between proposed method and other methods on the left atrium segmentation dataset.}
	\label{Fig2}
\end{figure*}

\begin{table*}[h]
	\caption{Comparative experimental results between our proposed method and other semi-supervised segmentation methods on LA dataset with 10\% annotation settings  (8 labeled scans and 72 unlabeled scans).} \label{Table2}
	\centering
	\renewcommand\arraystretch{1.5}
	\begin{tabular}{c|c|c|c|c|c}
		\hline 
		\textbf{Method}	& \textbf{Label/Unlabel}	& \textbf{Dice [\%]} & \textbf{Jaccard [\%]} & \textbf{ASD} & \textbf{95HD}  \\ \hline
		Supervised Baseline &  8/0              & 79.99 & 68.12 & 5.48 & 21.11    \\ \hline
        DAN \cite{zhang2017deep}  & 8/72                    &75.11   &63.47   &3.57   &19.04       \\
        CPS \cite{chen2021semi}  & 8/72                        &84.09    &73.17   &2.41    &22.55           \\
        Entropy Mini \cite{vu2019advent}  & 8/72             &85.90    &75.60   &2.74  &18.65                     \\
        DUW \cite{wang2020double}   & 8/72                      &85.91    &75.75   &3.31    &12.67              \\
        ICT \cite{verma2022interpolation}  & 8/72                        &85.39   &74.84    &2.88   &17.45                \\
        SASSNet \cite{li2020shape}  & 8/72                      &86.81   &76.92    &3.94   &12.54              \\
        DTC \cite{luo2021semi}   & 8/72                        &86.57    &76.55  &3.74   &14.47            \\
        CCT \cite{ouali2020semi}  & 8/72                       &85.49   &76.45   &2.99   &15.36          \\
        URPC \cite{luo2022semi}  & 8/72                     &85.02   &75.98   &2.95   &15.21               \\
        CPCL \cite{xu2022all}  & 8/72                       &86.08  &76.89   &2.33    &12.37             \\ \hline
        MT  \cite{tarvainen2017mean}   & 8/72             & 82.38 & 72.32  & 2.21 & 14.54    \\
	UA-MT  \cite{yu2019uncertainty}   & 8/72             & 84.25 & 73.48  & 3.36 & 13.84    \\ \hline
            \textbf{CE-MT(ours)} & 8/72        & \textbf{87.96} & \textbf{78.76} & \textbf{1.99} & \textbf{8.26}    \\ \hline 	
	\end{tabular}
\end{table*}

\section{Experiments}

\subsection{Datasets}

To evaluate the performance of proposed methods, we conduct experiments on the Left Atrium Segmentation Dataset \cite{xiong2021global} to evaluate the effectiveness of our proposed method.
The dataset contains 100 3D gadolinium-enhanced MR imaging scans (GE-MRIs) and corresponding LA segmentation mask for training and validation. These scans have an isotropic resolution of $0.625 \times 0.625 \times 0.625 mm^{3}$. According to the task setting in \cite{yu2019uncertainty}, we divide the 100 scans into 80 scans for training and 20 scans for testing, and apply the same pre-processing methods. Out of the 80 training scans, we use the same 20\%/16 scans as labeled data and the remaining 80\%/64 scans as unlabeled data for semi-supervised segmentation task.

\subsection{Implementation Details}

For implementation details of our experiments, we use V-Net \cite{milletari2016v} as the backbone structure with the same data split and implementation details as \cite{yu2019uncertainty} for all experiments to ensure a fair comparison. Our framework is implemented in PyTorch using an NVIDIA Tesla V100 GPU. 
We use the Stochastic Gradient Descent (SGD) optimizer to update the network parameters for 6000 iterations, with an initial learning rate 0.01 decayed by 0.1 every 2500 iterations. The batch size is 4, consisting of 2 labeled images and 2 unlabeled images. We randomly crop $112 \times 112 \times 80$ sub-volumes as the network input and  use a sliding window strategy to obtain the final segmentation.
During the training procedure, standard data augmentation techniques are applied to prevent over-fitting \cite{yu2017automatic}.

\subsection{Experimental Results}

To evaluate the effectiveness of our method, we conducted a comprehensive comparison with existing methods on the LA dataset.
Table \ref{Table0} presents the results of ablation experiments. Specifically, two different ensembling strategies named CE-MT-U and CE-MT-B in our framework is performed with comparison to classic self-ensembling \cite{tarvainen2017mean}.
Other than simple one-way ensembling from student model to teacher model, our proposed methods focus on ensembling from dual-task student models to teacher model under different competitive manner. 
From the experimental results, it can be observed that both two strategies outperform classic mean teacher model, demonstrating the effectiveness of the competitive ensembling strategy.
Compared with simply selecting out better student for ensembling in CE-MT-U, CE-MT-B seems to be a better selection of considering both two students for teacher models. However, the improvement is significant under the 16 labeled settings, while for 8 labeled setting, the improvement is limited. 
Overall, from the experimental results we can conclude that CE-MT-B is a better selection to utilize the information learned from both tasks.

Table \ref{Table1} show the comparison results of our proposed method with several recent state-of-the-art semi-supervised segmentation methods on under 20\% labeled (16 images) setting.
These methods include deep adversarial networks (DAN) \cite{zhang2017deep}, cross pseudo supervision (CPS) \cite{chen2021semi}, entropy minimization (Entropy Mini) \cite{vu2019advent}, double uncertainty weighted method (DUWM) \cite{wang2020double}, interpolation consistency training (ICT) \cite{verma2022interpolation}, shape-aware semi-supervised segmentation (SASS) \cite{li2020shape}, dual task consistency learning (DTC) \cite{luo2021semi}, cross-consistency training (CCT) \cite{ouali2020semi}, uncertainty rectified pyramid consistency (URPC) \cite{luo2022semi}, cyclic prototype consistency learning (CPCL) \cite{xu2022all}, mean teacher (MT) \cite{tarvainen2017mean}, and uncertainty-aware mean teacher (UA-MT) \cite{yu2019uncertainty}.
To ensure a fair comparison, we employed the same network backbone and training details for all of these methods.Our proposed method achieved significant performance gains, with the Dice score improving from 86.03\% to 89.78\% and the Jaccard score improving from 76.06\% to 81.62\%. In addition, Figure \ref{Fig2} provides visualizations of segmentation results from different semi-supervised segmentation methods. As observed, the segmentation results of our proposed method fit more accurately with the ground-truth masks, further verifying the effectiveness of our strategy.
Table \ref{Table2} show the comparison experimental results on 10\% labeled (8 images) setting.  Our proposed method can achieve significant performance gains, with the Dice score improving from 79.99\% to 87.96\% and the Jaccard score improving from 68.12\% to 78.76\%. Besides, our proposed method significantly outperforms other semi-supervised segmentation methods in terms of all the four evaluation metrics.

\section{Conclusion}

In this paper, we propose a novel competitive ensembling teacher student framework for semi-supervised for left atrium segmentation from 3D MR images. Specifically, two student models with different task-level disturbances are introduced to learn mutually from different perspectives. Instead of classic self-ensembling, we implement a competitive ensembling strategy to combine more reliable information to teacher model.
Our framework promotes the collaborative learning procedure of different student models under the guidance of teacher model and encourages different training networks to conduct a competitive learning and integration process. Ultimately leading to achieve better performance. We evaluate our proposed method on the public Left Atrium (LA) dataset.
Experimental results demonstrate substantial performance improvement by harnessing the potential of unlabeled data, as evidenced by the experimental results, with up to 3.75\% and 7.97\% in Dice and 5.56\% and 10.64\% in Jaccard coefficient compared to supervised baseline under 20\% labeled and 10\% labeled images setting, respectively. Additionally, our proposed method outperforms other semi-supervised segmentation methods with the same backbone network and task settings.
The proposed method exhibits strong potential for application in clinical settings, offering a possible solution to alleviate the burden of annotation costs. This has the potential to significantly cut down on the time effort, and resources required for annotating large amounts of medical data, making it a valuable tool in improving the efficiency and affordability of real-world clinical applications.

\bibliographystyle{IEEEtran}
\bibliography{ref}

\end{document}